\documentclass{article} 

\usepackage{hyperref}
\usepackage{url,color,graphicx}
\usepackage{sidecap}
\usepackage{booktabs}
\usepackage{multirow}

\usepackage{microtype}
\usepackage{graphicx}
\usepackage{subfigure}
\usepackage{booktabs} 
\usepackage{verbatim}

\newcommand{\mycomment}[1]{}

\newcommand{\fig}[1]{Fig.~\ref{fig:#1}}
\newcommand{\tab}[1]{Table~\ref{tab:#1}}
\newcommand{\secc}[1]{Section~\ref{sec:#1}}
\newcommand{\append}[1]{Appendix~\ref{app:#1}}

\usepackage[accepted]{icml2018}

\usepackage{algorithmicx}
\usepackage[noend]{algpseudocode}
\algdef{SE}[DOWHILE]{Do}{DoWhile}{\algorithmicdo}[1]{\algorithmicwhile\ #1}%

\icmltitlerunning{Composable Planning with Attributes}

\begin{document}
\twocolumn[
\icmltitle{Composable Planning with Attributes}



%
%
%

\icmlsetsymbol{equal}{*}

\begin{icmlauthorlist}
 \icmlauthor{Amy Zhang}{equal,fb}
  \icmlauthor{Adam Lerer}{equal,fb}
 \icmlauthor{Sainbayar Sukhbaatar}{nyu}
 \icmlauthor{Rob Fergus}{fb,nyu}
 \icmlauthor{Arthur Szlam}{fb}
 \end{icmlauthorlist}
 
 \icmlaffiliation{fb}{Facebook AI Research, New York, NY, USA}
 \icmlaffiliation{nyu}{New York University, New York, NY, USA}
 \icmlcorrespondingauthor{Adam Lerer}{alerer@fb.com}

\icmlkeywords{Machine Learning, ICML}

\vskip 0.3in
]

\printAffiliationsAndNotice{\icmlEqualContribution} 

\begin{abstract}
The tasks that an agent will need to solve often are not known during training. However, if the agent knows which properties of the environment are important then, after learning how its actions affect those properties, it may be able to use this knowledge to solve complex tasks without training specifically for them. Towards this end, we consider a setup in which an environment is augmented with a set of user defined attributes that parameterize the features of interest. We propose a method that learns a policy for transitioning between ``nearby'' sets of attributes, and maintains a graph of possible transitions. Given a task at test time that can be expressed in terms of a target set of attributes, and a current state, our model infers the attributes of the current state and searches over paths through attribute space to get a high level plan, and then uses its low level policy to execute the plan.  We show in 3D block stacking, grid-world games, and StarCraft that our model is able to generalize to longer, more complex tasks at test time by composing simpler learned policies.
\end{abstract}

\section{Introduction}

Researchers have demonstrated impressive successes in building agents that can achieve excellent performance in difficult tasks, e.g.~\cite{mnih2015,Silver2016}.   However, these successes have mostly been confined to situations where it is possible to train a large number of times on a single known task.   On the other hand,  in some situations, the tasks of interest are not known at training time or the space of tasks is so large that an agent will not realistically be able to train many times on any single task in the space.

  We might hope that the tasks of interest are {\it compositional}: for example, cracking an egg is the same whether one is making pancakes or an omelette.
If the space of tasks we want an agent to be able to solve has compositional structure, then a state abstraction that exposes this structure could be used both to specify instructions to the agent, and  to plan through sub-tasks that allow the agent to complete its instructions.

In this work we show how to train agents that can solve complex tasks by planning over a sequence of previously experienced simpler ones.   The training protocol relies on a state abstraction that is manually specified, consisting of a set of binary attributes designed to capture properties of the environment we consider important. These attributes, learned at train time from a set of (state, attribute) pairs, provide a natural way to specify tasks, and a natural state abstraction for planning.     Once the agent learns how its actions affect the environment in terms of the attribute representation, novel tasks can be solved compositionally by executing a plan consisting of a sequence of transitions between abstract states defined by those attributes.  Thus, as in  \cite{Dayan_FN,Dietterich_MAXQ,VezhnevetsOSHJS17}, temporal abstractions are explicitly linked with state abstractions.

Our approach is thus a form of model-based planning, where the agent first learns a model of its environment (the mapping from states to attributes, and the attribute transition graph), and then later uses that model for planning.  In particular, it is not a reinforcement learning approach, as there is no supervision or reward given for completing the tasks of interest.   Indeed,  outside of the (state, attribute) pairs, the agent receives no other reward or supervision.    In the experiments below, we will show empirically that this kind of approach can be useful on problems that can be challenging for standard reinforcement learning.

We evaluate compositional planning in several environments. We first consider 3D block stacking, and show that we can compose single-action tasks seen during training to perform multi-step tasks. Second, we plan over multi-step policies in 2-D grid world tasks. Finally, we see how our approach scales to a unit-building task in StarCraft.

\begin{figure*}[h!]
  \centering
\includegraphics[width=1.8\columnwidth]{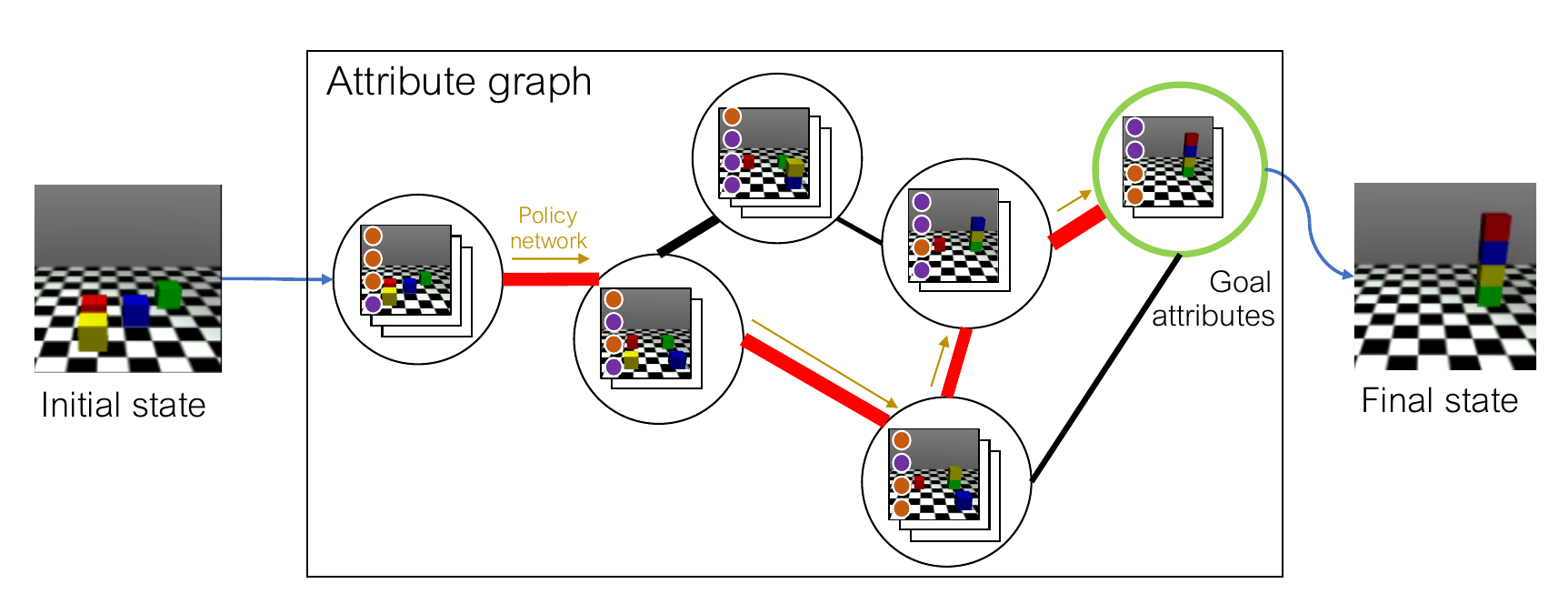}
\vspace{-4mm}
  \caption{\small Solving complex tasks by planning in attribute space. Each state is mapped to a set of binary attributes (orange/purple dots) which is learned from a set of labeled (state, attribute) pairs provided as input. One attribute might mean, for example, ``there is a blue block left of the red block''. The planning algorithm uses a graph over these sets of attributes with edge weights corresponding to the probability that a parametric policy network is able to transition from one attribute set to another. The graph and policy are learned during training via exploration of the environment. Given a goal attribute set (green), we use the graph to find the shortest path (red) to it in attribute space. The policy network then executes the actions at each stage (gold arrows).}
\label{fig:teaser}
\end{figure*}

\section{The \textit{Attribute Planner} Model}
\label{sec:model}

We consider an agent in a Markov environment, i.e. at each time the agent observes the state $s$ and takes action $a$, which uniquely determines the probability $P(s, a, s')$ of transitioning from $s$ to $s'$. We augment the environment with a map $f : S \to \{ \rho \}$ from states to a set of user-defined attributes $\rho$. We assume that either $f$ is provided or a small set of hand-labeled $(s, \rho)$ pairs are provided in order to learning a mapping $\hat{f}$. Hence, the attributes are human defined and constitute a form of supervision. Here we consider attributes that are sets of binary vectors. These user-specified attributes parameterize the set of goals that can be specified at test time.

The agent's objective at test time is, given a set of goal attributes $\rho_g$, to take a sequence of actions in the environment that end with the agent in a state that maps to $\rho_g$. During training, the agent constructs a model with three parts:

\begin{enumerate}
\item a neural-net based {\bf attribute detector} $\hat{f}$, which maps states $s$ to a set of attributes $\rho$, i.e. $\rho=\hat{f}(s)$.
\item a neural net-based {\bf policy $\pi(s,\rho_g)$} which takes a pair of inputs: the current state $s$ and attributes of an (intermediate) goal state $\rho_g$, and outputs a distribution over actions.
\item a {\bf transition table} $c_\pi(\rho_i, \rho_j)$ that records the empirical probability that $\pi(s_{\rho_i}, \rho_j)$ can succeed at transiting successfully from $\rho_i$ to $\rho_j$ in a small number of steps.
\end{enumerate}

The transition table $c_\pi$ can be interpreted as a graph where the edge weights are probabilities. This high-level attribute graph is then searched at test time to find a path to the goal with maximum probability of success, with the policy network performing the low-level actions to transition between adjacent attribute sets.

\subsection{Training the Attribute Planner}
\label{sec:training}

The first step in training the Attribute Planner is to fit the neural network detector $\hat{f}$ that maps states $s$ to attributes $\rho$, using the labeled states provided. If a hardcoded function $f$ is provided, then this step can be elided.

In the second step, the agent explores its environment using an exploratory policy. Every time an attribute transition $(\rho_i, \rho_j)$ is observed, it is recorded in an intermediate transition table $c_{\pi_e}$. This table will be used in later steps to keep track of which transitions are possible.

The most naive exploratory policy takes random actions, but the agent can explore more efficiently if it performs count-based exploration in attribute space. We use a neural network exploration policy that we train via reinforcement learning with a count-based reward proportional to $c_{\pi_e}(\rho_i, \rho_j)^{-0.5}$ upon every attribute transition $(\rho_i, \rho_j)$, where $c_{\pi_e}(\rho_i, \rho_j)$ is the visit count of this transition during exploration. This  bonus is similar to, for example, Model-based Interval Estimation with Exploration Bonuses \citep{strehl2008analysis}, but with no empirical reward from the environment. The precise choice of exploration bonus is discussed in Appendix $\ref{app:exploration}$.

Now that we have a graph of possible transitions, we next train the low-level goal-conditional policy $\pi$ and the main transition table $c_\pi$. From state $s$ with attributes $\rho$, the model picks an attribute set $\rho_g$ randomly from the neighbors of $\rho$ in $c_{\pi_e}$ weighted by their visit count in the Explore phase and sets that as the goal for $\pi$. Once the goal is achieved or a timeout is reached, the policy is updated and the main transition table $c_\pi$ is updated to reflect the success or failure. $\pi$ is updated via reinforcement learning, with a reward $r$ of $1$ if $\rho_g$ was reached and $0$ otherwise\footnote{Note that $c_\pi$ is collecting statistics about $\pi$ which is non-stationary. So $c_\pi$ should really be updated only after a burn-in period of $\pi$, or a moving average should be used for the statistics.}. See Algorithm \ref{alg:ap_training} for pseudocode of AP training.

In the case of block stacking (Sec. \ref{sec:stacking}), the attribute transitions consist of a single step, so we treat $\pi$ as an ``inverse model'' in the style of \cite{poking,AndrychowiczWRS17}, and rather than using reinforcement learning, we can train $\pi$ in a supervised fashion by taking random actions and training $\pi$ to predict the action taken given the initial state and final attributes.

\begin{algorithm}
\caption{Attribute Planner Training}
\label{alg:ap_training}

\begin{algorithmic}[t]
\State \textbf{Input:} Labeled pairs $\{(s_i, \rho_i)\}$, exploratory policy $\pi_e$, $N_1$, $N_2$, $t_{max}$.
\\
\State \textit{// Step 1: Train attribute detector}
\State Fit $\hat{f}$ on $\{(s_i, \rho_i)\}$ with supervised learning.
\\
\State \textit{// Step 2: Explore}
\For {$t = 1\ ...\ N_1$}
\State Act according to $\pi_e(s_{t-1})$.
\State Compute attributes $\rho_t \gets \hat{f}(s_t)$
\If {$\rho_t \neq \rho_{t-1}$}
\State Record the transition: $c_{\pi_e}(\rho_{t-1}, \rho_t) \mathrel{{+}{=}} 1$
\State Optional: Update $\pi_e$ with count-based reward.
\EndIf
\EndFor
\\
\State \textit{// Step 3: Train policy $\pi$ and $c_\pi$}
\State $t_{last} \gets 0, \rho_s \gets \emptyset , \rho_e \gets \textrm{RandNeighbor}(c_{\pi_e}, \rho_s)$
\For {$t = 1\ ...\ N_2$}
\State Compute attributes $\rho_t \gets \hat{f}(s_t)$
\If {$t=1$ or $\rho_t \neq \rho_s$ or $t - t_{last} \geq t_{max}$}
\State $r \gets 1$ if $\rho_t = \rho_e$, otherwise $0$.
\State UpdatePolicy($\pi, r$)
\State Record attempt: $A_\pi(\rho_{t-1}, \rho_t)\ \mathrel{{+}{=}} 1$
\State Record success: $S_\pi(\rho_{t-1}, \rho_t) \mathrel{{+}{=}} r$
\State $\rho_s \gets \rho_t, \rho_e \gets \textrm{RandNeighbor}(c_{\pi_e}, \rho_s)$
\State $t_{last} \gets t$
\EndIf
\State Take an action according to $\pi(s_{t-1}, \rho_e)$
\EndFor
\\
\For {$(\rho_i, \rho_j) \in A_\pi$}
\State $c_\pi(\rho_i, \rho_j) \gets S_\pi(\rho_i, \rho_j) / A_\pi(\rho_i, \rho_j)$.
\EndFor
\end{algorithmic}
\end{algorithm}

\subsection{Evaluating the model}

Once the model has been built we can use it for planning.  That is, given an input state $s$ and target set of attributes $\rho_T$, we find a path $[\rho_0, \rho_1, ... ,\rho_m]$ on the graph $G$ with $\rho_0 = f(s)$ and $\rho_m = \rho_T$ minimizing
\begin{equation}
\sum_{i=0}^{m-1} -\log c_\pi(\rho_i, \rho_{i+1})
\end{equation}
which maximizes the probability of success of the path (assuming independence). The probability $c_\pi$ is computed in Algorithm $\ref{alg:ap_training}$ as the ratio of observed successes and attempts during training.

The optimal path can be found using Dijkstra's algorithm with a distance metric of $-\log(c_\pi(\rho_i,\rho_{i+1}))$. The policy is then used to move along the resulting path between attribute set, i.e. we take actions according to $a  = \pi(s,\rho_1)$, then once $f(s)=\rho_1$, we change to  $a  = \pi(s,\rho_2)$ and so on.  At each intermediate step, if the current attributes don't match the attributes on the computed path, then a new path is computed using the current attributes as a starting point (or, equivalently, the whole path is recomputed at each step).

\begin{algorithm}
\caption{Attribute Planner Inference}
\begin{algorithmic}
\State \textbf{Input:} Low-level policy $\pi$, graph $c_\pi$, attribute detector $\hat{f}$, target attributes $\rho_T$.
\Do
\State $\rho \gets \hat{f}(s)$
\State $[\rho_0, ..., \rho_m ] \gets \textrm{ShortestPath}(c_\pi, \rho, \rho_g)$
\State Act according to $\pi(s_{t-1}, \rho_1)$.
\DoWhile{$\rho \neq \rho_T$}
\end{algorithmic}
\end{algorithm}

\subsection{An Aside: Which attributes should we include?}

Since we use user-specified attributes for planning, which attributes are important to include? The set of attributes must be able to specify our goals of interest, and should be parsimonious since extra attributes will increase the size of the graph and thus degrade the statistics on each edge/. 

On the other hand, the attributes should have a property that we will call ``ignorability'' which says that the probability of being able to transition from $\rho_i$ to $\rho_j$ should only depend on the attributes $\rho_i$, not the exact state; i.e. $P_\pi(f(s_{t'})=\rho_j | f(s_t)) = P_\pi(f(s_{t'})=\rho_j | s_t)$ \footnote{Note that the particular sequence of actions that effects the transition from $\rho_i$ to $\rho_j$ may still be conditional on the state.}.  To the extent that this condition is violated, then transitions are \textit{aliased}, and a planned transition may not be achievable by the policy from the particular state $s$ even though it's achievable from other states with the same properties, causing the model to fail to achieve its goal.  For example, in the block stacking task in \ref{sec:stacking}, there will be nontrivial aliasing; we will show that some amount of aliasing is not deadly for  our model.

\section{Related work}


\noindent{\bf Hierarchical RL}
Many researchers have recognized the importance of methods that can divide a MDP into subprocesses \citep{ThrunS94, Parr_HAM,Sutton99options, Dietterich_MAXQ}.   Perhaps the most standard formalism today is the options framework of \citep{Sutton99options},  which deals with multistep ``macro-actions'' in the setting of reinforcement learning.  Recent works, like \cite{KulkarniNST16}, have shown how options can be used with function approximation via deep learning.

Our work is also a hierarchical approach to controlling an agent in a Markovian environment.
However,  the paradigm we consider differs from reinforcement learning: we consider a setup where \emph{no reward or supervision is provided other than the $(s,f(s))$ pairs}, and show than an agent can learn to decompose a transition between far away $\rho, \rho'$ into a sequence of short transitions.  If we were to frame the problem as HRL,  considering each $\pi(\cdot,\rho)$ as a macro action\footnote{Note also that all the ``macro actions' in our examples in \ref{sec:stacking} are degenerate in the sense that they return after one step, but we still are able to show generalization to long trajectories from (unsupervised) training only on short ones}, in order for the agent to learn to sequence the $\pi(\cdot,\rho_i)$, the environment would need to give reward for the completion of complex tasks, not just simple ones.

As opposed to e.g. \cite{KulkarniNST16}, where additional human supervision is used to allow exploration in the face of extremely sparse rewards, our goal is to show that adding human supervision to parameterize the task space via attributes allows compositionality through planning.

In our experiments below, we compare our method to Option-Critic with a deliberation cost \cite{2017arXiv170904571H}, a hierarchical reinforcement learning method that learns options without additional supervision by including a  reward when a target goal is achieved.

\noindent{\bf Horde and descendants}
Our work is related to generalized value functions \citep{SuttonMDDPWP11} in that we have policies parameterized by state and target attributes. 
  Using attributes to parameterize the goal space  is related to the factored state-goal representation in \cite{Schaul2015universal}. In particular our work shares similarities with the formulation in  \cite{DosovitskiyK16}, which gives an agent supervision of future values of functions of the state considered important for describing tasks.  Unlike that work, our attributes are functions of the current state, and the model uses its own estimator to learn the dynamics at the level of attributes as a graph.  Thus, our model gets no extrinsic supervision of environment dynamics or goal attainment at the level of attributes. 
Finally,   \cite{Seijen_hybrid} used human provided attributes as a general value function (GVF) in Ms. Pacman, showing that using a weighted combination of these can lead to higher scores than standard rewards.  Although the representation used in that work is similar to the one we use, the motivation in our work is to allow generalization to more complicated tasks; and we use the attributes to guide exploration and to plan, rather than just as tools for building a reactive policy.

\noindent{\bf Factored MDP and Relational MDP}  Our approach is closely related to factored MDP \citep{Boutilier_exploiting,Boutilier_factored,GuestrinKPV03}.  In these works, it is assumed that the environment can be represented by discrete attributes, and that transitions between the attributes by an action can be modeled as a Bayesian network.   The value of each attribute after an action is postulated  to depend in a known way on attributes from before the action.   The present work differs from these in that the attributes do not determine the state and  the dependency graph is not assumed to be known.   More importantly, the focus in this work is on organizing the space of tasks through the attributes rather than being able to better plan a specific task; and in particular being able to generalize to new, more complex tasks at test time.

Our approach is also related to Relational MDP and Object Oriented MDP \citep{Hernandez-GardiolK03,Otterlo_Relational_RL, Diuk_object,AbelHBBOMT15}, where states are described as a set of objects, each of which is an  instantiation of canonical classes, and each instantiated object has a set of attributes.   Our work is especially related to \cite{GuestrinGeneralizing}, where the aim is to show that by using a relational representation of an MDP, a policy from one domain can generalize to a new domain.   However, in the current work, the attributes are taken directly as functions of the state, as opposed to defined for object classes, and we do not have any explicit encoding of how objects interact.  The model is given some examples of various attributes, and builds a parameterized model that maps into the attributes.

The Programmable Agents of \cite{DenilCCSF17}  put the notions of objects and attributes (as in relational MDP) into an end-to-end differentiable neural architecture. Their model also learns mappings from states to attributes, and is evaluated on a block manipulation task. In their work, the attributes are used to generalize to different combinations of object properties at test time, while we use it to generalize compositionally to more complex tasks. Also, while our model uses explicit search to reason over attributes, they use an end-to-end neural architecture.


\noindent{\bf Lifelong learning, multitask learning, and zero-shot learning}

There is a large literature on quickly adapting to a new learning problem given a set or a history of related learning problems.  Our approach in this work has a similar motivation to  \cite{IseleRE16}, where tasks are augmented with descriptors and featurized, and the coefficients of the task features in a sparse dictionary are used to weight a set of vectors defining the model for the associated task.   Similarly, the task is specified by a feature as an input into a model in \cite{Lopez-PazR17}.   However, in our work, although the attributes are used to parameterize tasks, rather than directly featurize the tasks, they are features of a state;  and we learn the mapping from state to attributes.    This allows our agent to learn how to transit between sets of attributes unsupervised, and plan in that space. 

Several recent deep reinforcement learning works have used modular architectures and hierarchy to achieve generalization to new tasks.   For example, \cite{Tessler_minecraft} uses pre-trained skills for transfer.   \cite{OhSLK17} uses a meta-controller that selects parameterized skills and analogical supervision on outer-product structured tasks.   
However, our ``meta-controller'' is the search over attributes, rather than a reactive model, which allows explicit planning.   Furthermore,   although our assignments of attributes serves a similar purpose to their analogical supervision (and outer-product task structure),
the methods are complementary; we can imagine augmenting our attributes with analogical supervision.

In \cite{AndreasKL17}, generalization is achieved through supervision in the form of ``policy sketches'', which are symbolic representations of the high level steps necessary to complete a given task.  The low level steps in executing modules in the sketches are composable.  Our work is similar in that high level annotation is used to enable generalization, but the mechanism in this work is different.   Note that the approaches in \cite{AndreasKL17} is also complementary to the one described here; in future work we wish to explore combining them.

\noindent{\bf Semiparametric methods}
In this work we use an explicit memory of sets of attributes the model has seen.  Several previous works have used non-parametric memories for lowering the sample complexity of learning, e.g. \cite{BlundellUPLRLRW16,PritzelUSBVHWB17}.   Like these, we lean on the fact that with a good representation of a state, it can be useful to memorize what to do in given situation (having only done it a small number of times)  and explicitly look it up.  In our case, the ``good representation'' is informed by the user-specified attributes.

Our approach is also related to \cite{MachadoBB17}, which builds up a multiscale representation of an MDP using Eigenvectors of the transition matrix of the MDP, in the sense that we collect data on possible transitions between attributes in a first phase of training, and then use this knowledge at test time.

\noindent{\bf Learning symbolic representations for planning}  There is a large literature on using symbolic representations for planning, for example the STRIPS formalism \cite{STRIPS}.  In  \cite{KonidarisKL18}, the authors  propose a model that learns the symbols for a STRIPS-style representation.   Like in our work, their model learns the interface between the raw state observations and the planner.   However, in that work, the abstract structure is given by a set of pre-defined options with fixed policies.


\section{Experiments}

We perform experiments with the Attribute Planner (AP) in three environments. First, we consider a 3D block stacking environment. Here, we demonstrate that AP allows compositional generalization by training a low level policy on single-action tasks in a supervised fashion and showing that with the AP algorithm it can perform multi-step tasks at test time.

Second, we consider 2D grid worlds in Mazebase \citep{sukhbaatar2015mazebase}, where we evaluate AP's performance when the low-level policy is temporally extended and must be learned via RL.

Finally, we evaluate AP on a build order planning task in Starcraft to see how AP scales to a more complex task that is of broader interest We further show that an exploratory policy over attributes allows the agent to explore attribute transitions where random search fails.

\textbf{Baselines:} In all experiments, we compare against baseline policies trained with reinforcement learning. These baseline policies take the state and goal as inputs, and use the same neural network architecture as the policy used for the Attribute Planner. We consider several training regimes for the baseline policies: (i) training only with nearby goals like AP; (ii) training on the multi-step evaluation tasks; and (iii) training on a curriculum that transitions from nearby goals to evaluation tasks. Policies (ii) and (iii) are trained on full sequences, thus have an inherent advantage over our model.

In the block stacking task, we further compare against a state-of-the-art algorithm for hierarchical RL: Option-Critic with deliberation cost \cite{2017arXiv170904571H}, as well as an inverse model trained by supervised learning.

\subsection{Block Stacking}
\label{sec:stacking}

We consider a 3D block stacking environment in Mujoco \cite{TodorovET12}. In this experiment, \textit{we train AP only on single-action trajectories} and evaluate on multi-step tasks, in order to evaluate AP's ability to generalize using planning. We compare AP with baselines trained on both single-action, multi-action, and curriculum tasks.

In this environment, there are $4$ blocks of different colors, and actions consist of dropping a block in a $3\times 3$ grid of positions, resulting in $36$ total actions. A block cannot be moved when it is underneath another block, so some actions have no effect.

The input to the model is the observed image, and there are a total of $36$ binary properties corresponding to the relative x and y positions of the blocks and whether blocks are stacked on one another. For example, one property corresponds to ``blue is on top of yellow''. Each training episode is initiated from a random initial state and lasts only one step, i.e. dropping a single block in a new location. Further model and training details and results on a continuous variant of this environment are provided in Appendix \ref{app:blocks}.

Table \ref{tab:blocks_main} compares the performance of different models on several block stacking tasks. 
In the \textit{multi-step} task, the goal is chosen as the properties of a new random initialization. These tasks typically require $3 - 8$ steps to complete. In the \textit{4-stack} task, the goal is a vertical stack of blocks in the order red, green, blue, yellow. In the \textit{underspecified} task, we consider a multi-step goal where only 70\% of the attributes are provided at random. The AP model handles these naturally by finding the shorted path to any satisfactory attribute set.

The single-step reactive policies perform similarly to AP when evaluated on the single-step tasks it sees during training (see Table \ref{tab:blocks_main_full} in the Appendix), but perform poorly when transferred to multi-step tasks, while AP generalizes well to complex task. The AP model also solves underspecified tasks even though they are not seen explicitly during training.

The attribute detector $\hat{f}$ predicts the full attribute set with $<0.1\%$ error when trained on the full dataset of 1 million examples. If trained on only 10,000 examples, the attribute detector has an error rate of $1.4\%$. Training the AP model with this less-accurate attribute detector degrades multi-step performance by only 0.9\%.

\begin{table}[h]
\small
\begin{center}
\begin{tabular}{ l l | c c c  c c }
\toprule
Model          & Training  & multi-step  & 4-stack & underspec. \\
               & Data      & \%          &    \%   & \% \\
\midrule
A3C            & 1S        & 8.1         & 1.9     & 6.6     \\
A3C            & MS        & 0.0         & 0.0     & 0.0     \\
A3C            & C         & 17.0        & 2.9     & 0.2     \\
Option-Critic  & 1S        & 0.6         & 1.0     & 1.2     \\
Option-Critic  & MS        & 0.2         & 0.5     & 1.7     \\
Option-Critic  & C         & 0.4         & 0.9     & 1.0     \\
Inverse        & 1S        & 9.1         & 0.5     & 18.8    \\  
Inverse        & MS        & 13.7        & 4.6     & 9.6     \\
\midrule
AP (no $c_\pi$)& 1S        & 29.7        & 62.2    & 28.1    \\
AP             & 1S        &\bf{66.7}    &\bf{98.5}&\bf{63.5}\\
\bottomrule
\end{tabular}
\end{center}
\caption{Key: 1S = one-step; MS = multi-step; C = curriculum. Model comparison on block stacking task accuracy. Baselines marked `multi-step' or `curriculum` get to see complex multi-step tasks at train time.  The Attribute Planner (AP) generalizes from one-step training to multi-step and underspecified tasks with high accuracy, while reinforcement learning and inverse model training do not. AP outperforms baselines even with a curriculum of tasks. Ablating the normalized graph transition table $c_\pi$ degrades AP performance substantially on multi-step tasks due to aliasing. }
\label{tab:blocks_main}
\end{table}

We also consider a variant of the block stacking task with a continuous action space, in which an action consists of dropping a block at any x-y position. While performance degrades substantially for all models in the continuous action space, AP continues to outperform reactive policies on multi-step tasks. See Appendix \ref{app:blocks} for the full results.

\begin{figure}[h]
\begin{center}
\includegraphics[width=\columnwidth]{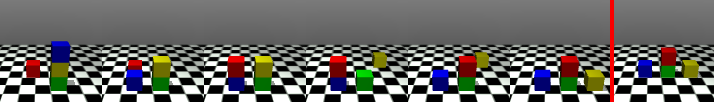}\\
\includegraphics[width=\columnwidth]{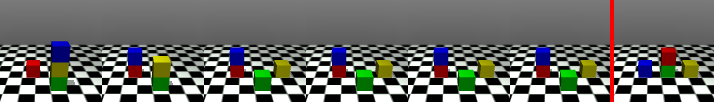}\\
\vspace{-3mm}
\rule{\columnwidth}{1.5pt}
\includegraphics[width=\columnwidth]{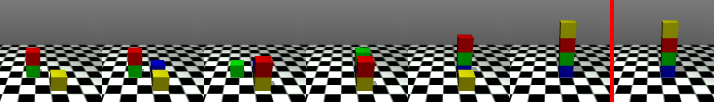}\\
\includegraphics[width=\columnwidth]{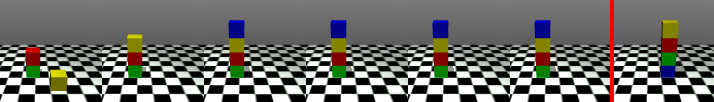}
\vspace{-6mm}
\caption{Two examples of block stacking evaluation tasks. The initial/target states are shown in the first/last columns. Successful completions of our Attribute Planner model are shown in rows 1 and 3. By contrast, the A3C baseline is unable to perform the tasks (rows 2 and 4).}
\label{fig:blocks_example}
\end{center}
\end{figure}
\begin{figure}[h]
\begin{center}
\vspace{-1mm}
\includegraphics[width=\columnwidth]{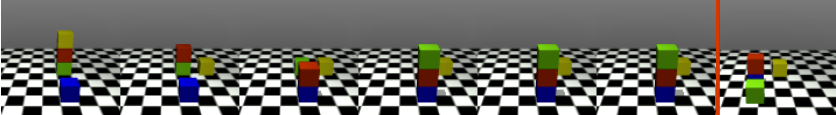}
\vspace{-1mm}
\caption{Plans become stuck when states with different transitions map to the same properties. In frame 4 of this example, the policy is directed to place the green block in front of the red and blue blocks, but this is impossible because the blue and red are already in the frontmost position.}
\vspace{-3mm}
\label{fig:blocks_aliasing}
\end{center}
\end{figure}
\setlength\tabcolsep{4pt} 
\begin{table}[h]
\small
\begin{center}
\begin{tabular}{ l | c c | c c c}
\toprule
\# Training & \multicolumn{2}{c}{Inverse}   & \multicolumn{3}{c}{AP} \\
\ Examples & 1-step & multi-step & 1-step & multi-step & \# edges \\
\midrule
10,000 & 35.5\% & 1.6\% &  50.0\% & 3.0\% & 11438\\
100,000 & 99.9\% & 7.8\% & 89.0\% & 47.0\% & 57299\\
1,000,000 & 100\% & 9.1\% & 98.9\% & 66.7\% & 144083 \\
10,000,000 & 100\% & 8.5\% & 96.5\% & 70.7\% & 238969\\
\bottomrule
\end{tabular}
\end{center}
\caption{Effect of the number of (one-step) training examples on one-step and multi-step task performance, for an inverse model and the Attribute Planner model. The AP model does slightly worse on 1-step tasks than an inverse model for large numbers of training examples, but generalizes to multi-step tasks. The number of edges continues to increase even after 1 million examples, but these extra edges do not make a big difference for planning because there are multiple paths to a given goal. }
\label{tab:blocks_sample}
\end{table}
\setlength\tabcolsep{6pt} 

\noindent {\bf Property Aliasing:} The ``ignorability'' assumption we made in Section \ref{sec:model} is violated in the block stacking task. To see why, consider a transition from ``red left of blue and yellow" to ``red right of blue and yellow". This can typically be accomplished in one step, but if blue and yellow are already on the far right, it cannot. Thus, states where this transition are possible and impossible are aliased with the same properties. Table \ref{tab:blocks_sample} shows that the performance is nearly perfect for individual transitions (1-step tasks), and the graph is well-connected after 1 million training examples, so the main source of error on these tasks is in fact aliasing. Figure \ref{fig:blocks_aliasing} shows an example plan that becomes stuck due to aliasing.

The transition table $c_\pi$ is important for mitigating the effects of aliasing in the block stacking task. The graph search finds the path with the highest probability of success (i.e. the product of probabilities on each edge), so it avoids edges that have high aliasing. In Table \ref{tab:blocks_main}, we consider an ablation of $c_\pi$ from the AP model, in which the probability of transitioning from an edge $(\rho_i, \rho_j)$ is estimated as the fraction of transitions from $\rho_i$ that ended in $\rho_j$ during the Explore phase. This ablation performs substantially worse than the full AP model.

\subsection{Grid Worlds}
\label{sec:mazebase}

We next consider tasks in which a multi-step low-level policy is required to transition between neighboring attributes.

We consider two classes of small 2-$D$ environments in Mazebase~\citep{sukhbaatar2015mazebase}, where the worlds are randomly generated for each episode. The action space for each consists of movements in the four cardinal directions plus additional environment-specific actions.

\begin{figure}[h]
\begin{center}
\includegraphics[width=1.0\columnwidth]{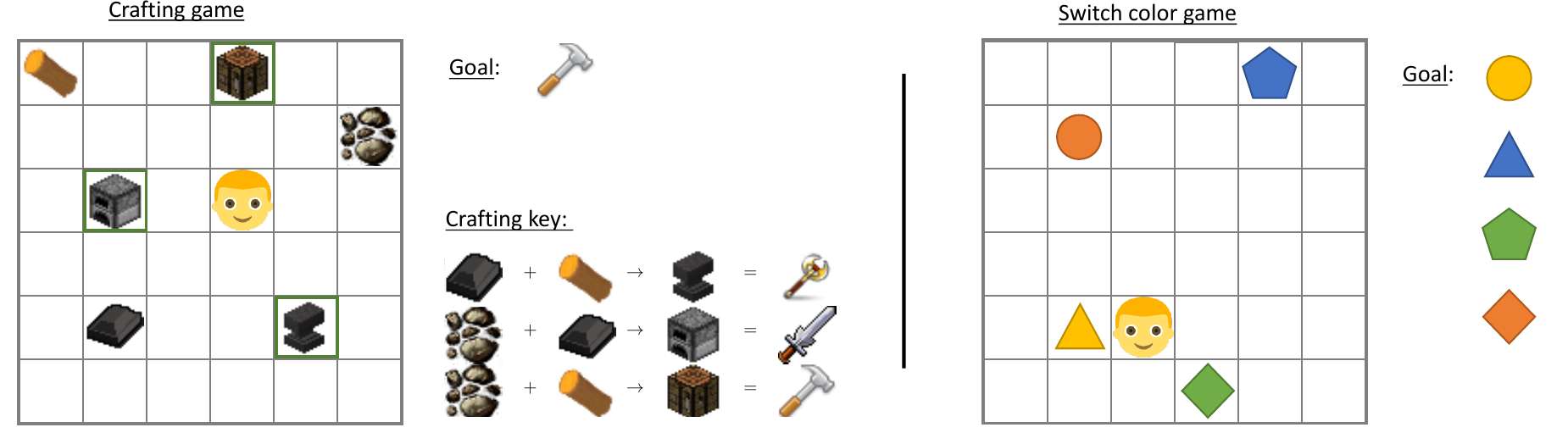}
\vspace{-4mm}
\caption{Left: Crafting mazebase game. Right: Colored switches game. See text for details.}
\label{fig:mazebase}
\vspace{-4mm}
\end{center}
\end{figure}

\noindent {\bf Colored Switches}
The first environment consists of four switches, each with four possible colors. An extra toggle action cycles the color of a switch if the agent is standing on it. The attributes for this environment are the states of the switches and the tasks are to change the switches into a specified configuration, as shown in \fig{mazebase}(right). The locations and colors of the switches are randomly initialized for each episode.

\noindent {\bf Crafting} In the second environment, similar to the one used in \cite{AndreasKL17},  an agent needs to collect resources and combine them to form items.  In addition to moving in the cardinal directions, the agent has a  ``grab''  action that allows it to pick up a resource from the current location and add it to its inventory.  The agent also has a ``craft''  action that combines a set of items to create a new item if  the agent has the prerequisite items in its inventory and  the agent is standing on a special square (a ``crafting table'')  corresponding to the item to be crafted. The attributes for this environment are the items in the inventory, and the task is to add a specified (crafted) item to the inventory. In the environment, there are three types of resources and three types of products  (see \fig{mazebase}(left)). The game always starts with three resources and an empty inventory.

In both environments, the agent's observation consists of a bag of words, where the words correspond to (feature, location). Features consist of item types, names, and their other properties. The locations include position relative to the agent in the maze, and also a few special slots for inventory, current, and target attributes.

Training proceeds according to Algorithm \ref{alg:ap_training}. During the explore phase, an exploratory policy is trained with reinforcement learning using a count-based reward proportional to $$\left( \frac{c_{\pi_e}(\rho_i,\rho_j) }{ \sum_{i,j}{c_{\pi_e}(\rho_i,\rho_j)}} + 0.001 \right)^{-0.5}$$ for making a transition $(\rho_i, \rho_j)$, where $c_{\pi_e}(\rho_i,\rho_j)$ is the number of times transition $(\rho_i,\rho_j)$ has been seen so far. We discuss this exploration bonus in Appendix \ref{app:exploration}. 

During the final phase of training we simultaneously compute $\pi$ and $c_\pi$, so we use an exponentially decaying average of the success rate of $\pi$ to deal with it's nonstationarity:
\[ c_\pi(\rho_i, \rho_j) = \frac{\sum_{t=1}^T \gamma^{T-t} S_\pi^t(\rho_i, \rho_j)}{\sum_{t=1}^T \gamma^{T-t} A_\pi^t(\rho_i, \rho_j)}, \]
where $T$ is the number of training epochs, $A_\pi^t$ is the number of attempted transitions $(\rho_i, \rho_j)$ during epoch $t$, and $S_\pi^t$ is the number of successful transitions. A decay rate of $\gamma=0.9$ is used. More details of the model and training are provided in Appendix \ref{app:mazebase}.

In the switches environment, multi-step test tasks are generated by setting a random attribute as target, which can require up to 12 attribute transitions. In the crafting environment, test tasks are generated by randomly selecting a (crafted) item as a target.   Since we do not care about other items in the inventory, the target state is underspecified.

We produce a curriculum baseline by gradually increase the upper bound on the difficulty of tasks during training. In the switches environment, the difficulty corresponds to the number of toggles necessary for solving the task. The craft environment has two levels of difficulty: tasks can be completed by a single grab or craft action, and tasks that require multiple such actions.


\begin{table}[h!]
  \centering
  \begin{tabular}{l l | c c c c }
    \toprule
    \multirow{ 2}{*}{Method} &Training& Switches & Crafting \\
    &Tasks & \% & \% \\
    \midrule
    Reinforce & 1-step       & 15.4  & 26.0  \\
    Reinforce & test tasks  &  0.0  & 21.8  \\ 
    Reinforce & curriculum & 33.3 & 51.5  \\ 
    AP            &  -               &\bf{83.1} &\bf{99.8}  \\
    \bottomrule
  \end{tabular}
  \caption{Task success rate on Mazebase environments. The Attribute Planner (AP) outperforms reactive policies trained with the Reinforce algorithm on multi-step evaluation tasks, even if the reactive policies are trained with a curriculum of tasks. "Training tasks" for AP are not specified because its unsupervised learning does not see tasks at training time.}
\label{tab:mazebase}
\end{table}

Table \ref{tab:mazebase} compares our Attribute Planner (AP) to a reinforcement learning baseline on the mazebase tasks. The AP planning outperforms purely reactive training regardless of whether one-step, multi-step, or a curriculum of training examples is provided. 

\subsection{StarCraft}
Finally, we test our approach for planning a build order in StarCraft: Brood War~\cite{synnaeve2016torchcraft}.
We consider the space of tasks of building particular units in a fixed time of 500 steps, e.g. ``build 1 barracks and 2 marines``.

This task is challenging for RL because the agent must complete a number of distinct steps, e.g. mine enough ore, then build a barracks, and finally train marines using the barracks, before receiving a reward. Each of these steps requires the agent have to control multiple units of different types using low-level actions similar to how a human plays the game. See \append{sc} for more details.

As in~\cite{SukhbaatarKSF17}, we restrict the game to the Terran race and only allow construction of certain units. In the small version, the agent can mine ore and build SCVs, supply depots, barracks, and marines. In the large version, an engineering bay and missile turrets are included as well. The attributes are chosen to be the number of units and resources of each type, specifically
\[ \{
\min(\lfloor N_{ore}/25 \rfloor , 40), N_{SCV}, N_{depot}, N_{barracks}, N_{marine}
\}\]
where $N_x$ is the number of $x$ present in the game, including units under construction. The large version also include $\{ N_{eng.bay}, N_{turrets} \}$.

Models are trained for a total of 30 million steps. AP uses 16 million steps for exploration and 14 million steps for training $\pi$.

\tab{starcraft} shows the final performance of the AP model and reactive RL baselines on this task after 30 million steps of training. \begin{table}[h!]
  \centering
  \begin{tabular}{l l | c c }
    \toprule
    \multirow{ 2}{*}{Method} & Training & \textit{Small} & \textit{Large} \\
     & Tasks & \textit{version} & \textit{version} \\
    \midrule
    Reinforce & test tasks  & 12.6\% & 2.3\%  \\ 
    Reinforce & curriculum & 18.9\% & 1.9\% \\ 
    AP  & -   & \bf{31.7\%} & \bf{35.2\%} \\
    \bottomrule
  \end{tabular}
  \caption{Comparison of AP with a policy trained with reinforcement learning on Starcraft building tasks. Unlike RL, the AP does not see tasks during training. Reported numbers are an average of 5 independent runs with different random seeds. AP outperforms reinforcement learning even when a curriculum is provided. AP scales dramatically better than RL as the task becomes more complex, because it can perform smart exploration in attribute space and can plan over a set of simple tasks that do not grow in complexity.}
\label{tab:starcraft}
\end{table}

AP exploration finds 120,000 and 420,000 edges for the small and large versions, respectively. The size and scaling of this graph show the limitations of a fully explicit graph. In fact, we represent the ore attribute as $\lfloor N_{ore}/25 \rfloor$ because it decreases the size of the graph by a factor of 25: otherwise for each transition w.r.t. the other attributes, the graph would have a separate edge for each valid value of total ore.

Count-based exploration over attributes is vital during the Explore phase in StarCraft. If a random policy is used in the small version, only 2047 edges are discovered as opposed to 120,000 using count-based exploration, and the final performance is reduced from 31.7\% to 6.4\%.


\section{Discussion}
Our results show that structuring the space of tasks with high level attributes allows an agent to compose policies for simple tasks into solutions of more complex tasks.   The agent plans a path to the final goal at the level of the attributes, and executes the steps in this path with a reactive policy.  Thus, supervision of an agent by labeling attributes can lead to generalization from simple tasks at train time to more complex tasks at test time. There are several fronts for further work:

\noindent {\bf Sample complexity of the planning module:}
In Table \ref{tab:blocks_sample} we can see both the benefits and the liabilities of the explicit non-parametric form for $c_\pi$.  By 10K samples, the parametric lower level policy is already able to have a reasonable success rate.  However, because in this environment, there are over 200K edges in the graph, most of the edges have not been seen, and without any weight-sharing, our model cannot estimate these transition probabilities.  On the other hand, by 100K samples the model has seen enough of the graph to make nontrivial plans; and the non-parametric form of the graph makes planning straightforward.   

In future work, we hope to combine parametric models for $c_\pi$ with search to increase the sample efficiency of the planning module.    Alternatively, we might hope to make progress on dynamic abstraction (projecting out some of the attributes) depending on the current state and goal, which would make the effective number of edges of the graph smaller.

\noindent {\bf Exploration} We have shown that the attributes $\rho$ and counts $c_\pi$, in addition to their usefulness for planning, provide a framework for incentivizing exploration. In this work we considered a simple count-based exploration strategy, which achieved better exploration in attribute space than random exploration. However, this setting of \textit{pure exploration} where there are no empirical rewards is different from the classic problem of exploration in an MDP, and warrants further exploration (see Appendix \ref{app:exploration}).

\noindent {\bf Learning the attributes:} Discovering the attributes automatically would remove much of the need for human supervision. Recent work, such as \cite{Thomas17}, demonstrates how this could be done.  Another avenue for discovering attributes is to use a few ``seed'' attributes, which is necessary for task specification anyway, and use aliasing as a signal that some attributes need to be refined.


\bibliography{boxer_arxiv.bbl}
\bibliographystyle{iclr2018_conference}

\appendix
\newpage

\section{Exploration Over Attributes}

\label{app:exploration}
Traditional count-based exploration adds a reward bonus proportional to $N^{-0.5}$ in a reinforcement learning context where an empirical reward is provided. For finite MDPs, this bonus decays to 0 as $t\to \infty$, which means that in the long-time limit the agent is still finding a policy that optimizes the original MDP.

In our setting, however, we are interested in \textit{pure exploration}, where the goal is to sample states (attribute transitions) somewhat uniformly, or to minimize the uncertainty in the empirical transition probabilities $c_\pi$. In this setting, it's problematic for the reward to converge to $0$ uniformly as $t \to \infty$, because the optimal policy becomes degenerate when $r_t \to 0$.

Instead, we consider a reward of $r_t = f(t/N)$, where $N$ is the visit count to this state (states are $(\rho_i, \rho_j)$ transitions in our example). If $f$ is a concave increasing function (e.g. $f(x) = x^{0.5}$) then in a bandit setting$$\lim_{T\to \infty} {\sum_{t=0}^T r_t/T}$$ is maximized when states are visited uniformly. In an MDP setting where you can't achieve exactly uniform exploration, the maximum of this reward will be determined by the choice of $f$.

For Mazebase experiments (with a small graph), we found that the standard count-based reward of $N^{-0.5}$ actually performed worse than a random policy, while a reward of $$\left( \frac{ N }{ t } + 0.001 \right)^{-0.5}$$ (which is a smoothed version of $(t/N)^{0.5}$ to prevent large rewards which can destabilize training) outperformed random exploration. In the crafting environment, a random agent finds 18.6 edges on average, while the exploratory policy finds all 25 edges. 

In StarCraft, which had a much larger graph with at least 400K edges, we found that $N^-0.5$ actually worked fine for exploration, discovering more than 50x more edges than a random policy.

\section{Details of the Block Stacking experiment}
\label{app:blocks}

The policy network $\pi$ takes (i) a $128\times128$ image, which is featurized by a CNN with five convolutional layers and one fully connected (fc) layer to produce a 128d vector; and (ii) goal properties expressed as a 48d binary vector, which are transformed to a 128d vector. The two 128d vectors are concatenated and combined by two fc layers followed by softmax to produce an output distribution over actions. We use an exponential linear nonlinearity after each layer.

For the underspecified task, we consider the same distribution of tasks as the multi-step task, but provide only 70\% of the attributes as the goal, at random. For the baseline models, we train them with the same distribution of underspecified goals as they see at test time, but this is not necessary for the AP model since it can plan over all goals that satisfy the underspecified goal.

Tables \ref{tab:blocks_main_full} and \ref{tab:blocks_main_cont_full} show the performance of AP and baselines on both one-step and multi-step evaluation tasks, in $3 \times 3$ and continuous action spaces, respectively. In the continuous case, blocks can be dropped in any $(x, y)$ with a fixed height $z=1.5$ and $x, y\in [0, 1]$.  The action space consists of choosing a discrete block id $b_{id} \in [0, 1, 2, 3]$ and continuous $x, y$. The inverse one-step models were trained on 2 million examples, inverse multi-step and AP models were trained on 1 million examples, and A3C models were trained to convergence (approximately 10 million examples). We perform each evaluation task 1000 times.

\begin{table*}[h]
\begin{center}
\begin{tabular}{ ll | c c c  c c }
\toprule
Model          & Training     & 1-step     & multi-step      & 4-stack & 1-step          & multi-step \\
               & Data         & \%         & \%      &    \%   & \multicolumn{2}{c}{underspecified} \\
\midrule
A3C            & 1-step        & 98.5   & 8.1     & 1.9     & 65.7    & 6.6     \\
A3C            & multi-step    & 2.6    & 0.0     & 0.0     & 5.3     & 0.0     \\
A3C            & curriculum    & 98.2   & 17.0    & 2.9     & 8.2     & 0.2     \\
Option-Critic  & 1-step        & 33.3   & 0.6     & 1.0     & 34.9    & 1.2     \\
Option-Critic  & multi-step    & 15.9   & 0.2     & 0.5     & 32.9    & 1.7     \\
Option-Critic  & curriculum    & 32.7   & 0.4     & 0.9     & 32      & 1.0     \\
Inverse        & 1-step        &\bf{100}& 9.1     & 0.5     &\bf{98.8}& 18.8    \\  
Inverse        & multi-step    & 94.1   & 13.7    & 4.6     & 71.2    & 9.6     \\
\midrule
AP (no $c_\pi$)& 1-step        & 74.5   & 29.7    & 62.2    & 81.8    & 28.1    \\
AP             & 1-step        & 98.8   &\bf{66.7}&\bf{98.5}& 97.8    &\bf{63.5}\\
\bottomrule
\end{tabular}
\end{center}
\caption{Model comparison on block stacking task accuracy. Baselines marked `multi-step' or `curriculum` get to see complex multi-step tasks at train time.  The Attribute Planner (AP) generalizes from one-step training to multi-step and underspecified tasks with high accuracy, while reinforcement learning and inverse model training do not. AP outperforms baselines even with a curriculum of tasks. Ablating the normalized graph transition table $c_\pi$ degrades AP performance substantially on multi-step tasks due to aliasing. Inverse one-step model was trained on 2 million examples, inverse multi-step and AP models were trained on 1 million examples, A3C models were trained to convergence.}
\label{tab:blocks_main_full}
\end{table*}

\begin{table*}[h]
\begin{center}
\begin{tabular}{ l l | c c c  c c }
\toprule
Model          & Training  & 1-step      & multi-step      & 4-stack & 1-step          & multi-step \\
               & Data      & \%      & \%      &    \%   & \multicolumn{2}{c}{underspecified} \\
\midrule
A3C            & 1-step        &  0.3    & 0.0     & 0.0     & 0.0     & 0.0     \\
A3C            & multi-step    &  0.0    & 0.0     & 0.0     & 0.5     & 0.0     \\
A3C            & curriculum    &  0.3    & 0.0     & 0.0     & 0.0     & 0.0     \\
Option-Critic  & 1-step        & 13.8    & 0.0     &\bf{4.4} & 14.6    & 0.0     \\
Option-Critic  & multi-step    & 14.4    & 0.0     & 0.0     & 13.1    & 0.0     \\
Option-Critic  & curriculum    & 15.2    & 0.3     & 1.3     & 15.2    & 0.1     \\
Inverse        & 1-step        & 60.1    & 8.6     & 0.0     & 29.3    & 2.5     \\  
Inverse        & multi-step    & 0.7     & 0.0     & 0.0     & 1.0     & 0.1     \\
\midrule
AP (no $c_\pi$)& 1-step        & 56.6    & 13.6    & 0.0     & 40.0    & 10.3    \\
AP             & 1-step        &\bf{64.1}&\bf{17}  & 0.4     &\bf{45.3}&\bf{12.5}\\
\bottomrule
\end{tabular}
\end{center}
\caption{Model comparison in the \textit{continuous} block stacking task. Inverse one-step model was trained on 8 million examples, inverse multi-step and AP models were trained on 10 million examples, A3C models were trained to convergence.}
\label{tab:blocks_main_cont_full}
\end{table*}

\section{Details of the Mazebase experiment}
\label{app:mazebase}
The policy network is a fully connected network with two hidden layers of 100 units.  The policy is trained with the Reinforce algorithm \citep{Williams92simplestatistical} to reach a neighboring set of attributes from the current state. Each round of training episodes terminate when the task completes or after 80 steps (i.e. $t_{max}=80$); and a reward of 1 is received if the goal is reached\footnote{Note that once the attribute detectors are learned, the reward is intrinsic: the agent considers a local task complete when {\it it} decides the attributes have changed appropriately}. An additional reward of -0.1 is given at every step to encourage the agent to complete the task quickly. We run each experiment three times with different random seeds, and report the mean success rate.

Some crafting tasks are pre-solved because the randomly chosen target item can already be in the inventory. However, such tasks have no effect on training and are also removed during testing.

\section{Details of the StarCraft experiment}
\label{app:sc}

The game initializes with 4 SCVs, a command center, and nearby ore mines.
The other types of units can be constructed, such as barracks, marines, supple depots, engineering bays, and missile turrets.
However, the policy only controls SCVs, a command center, and barracks.
The available actions to the policy differ depending on the unit type:
\begin{itemize}
    \item {\bf SCVs:} movements in 4 cardinal directions, mine ore, build a barracks, build a supple depot (also build a engineering bay, build a missile turret in the large version)
    \item {\bf Command center:} train a SCV
    \item {\bf Barracks:} train a marine
\end{itemize}
Each units observes its 64x64 surrounding area with resolution of 4.
Every time step, a policy outputs an action for each unit independently by taking its observation and the current attributes as an input.

Although the exact placement of units can be of importance in the game, here we only focus on their count.
Hence, the attributes are chosen to be the number of units and resources of each type.
Since any single unit can't observe everything in the game, detecting attributes from the observation alone is impossible. Therefore, the attributes are given as a part of the observation.

Multi-step tasks are generated by picking a random number for each unit type, with exception of ore and SCVs.
The upper limits of those random number are set between 2 and 4 depending on the unit type.

The same reinforcement training procedure as \secc{mazebase} is employed for the RL baselines.
For curriculum training, the upper limits are linearly increased during the curriculum training to make learning easy. Both baselines are trained for 30M steps.

Each episode starts at the initial state of the game and lasts 500 steps.
The exploration policy $\pi_e$ is trained with reinforcement learning with an intrinsic reward similar to Mazebase, although we find that scaling by number of transitions is unnecessary so we just use $c_{\pi_e}(\rho_i, \rho_j)^{-0.5}$ at each transition is more effective in StarCraft.  The exploration policy is trained for 16 million steps, followed by training $\pi$ and $c_\pi$ for 14 million steps, with $t_{max}=50$.

%
%
%

\end{document}